# Improved Mean and Variance Approximations for Belief Net Responses via Network Doubling

**Peter Hooper**  
Dept of Mathematical & Statistical Sciences  
University of Alberta  
Edmonton, AB T6G 2G1 Canada  
hooper@stat.ualberta.ca

**Yasin Abbasi-Yadkori, Russ Greiner, Bret Hoehn**  
Dept of Computing Science  
University of Alberta  
Edmonton, AB T6G 2H1  
{abbasiya, greiner, hoehn}@cs.ualberta.ca

## Abstract

A Bayesian belief network models a joint distribution with an directed acyclic graph representing dependencies among variables and network parameters characterizing conditional distributions. The parameters are viewed as random variables to quantify uncertainty about their values. Belief nets are used to compute responses to queries; i.e., conditional probabilities of interest. A query is a function of the parameters, hence a random variable. Van Allen *et al.* (2001, 2008) showed how to quantify uncertainty about a query via a delta method approximation of its variance. We develop more accurate approximations for both query mean and variance. The key idea is to extend the query mean approximation to a "doubled network" involving two independent replicates. Our method assumes complete data and can be applied to discrete, continuous, and hybrid networks (provided discrete variables have only discrete parents). We analyze several improvements, and provide empirical studies to demonstrate their effectiveness.

## 1 INTRODUCTION

Consider a simple example. Suppose $A$ represents presence/absence of a medical condition while $B$ and $Y$ are test results. Variables $B$ and $Y$ are conditionally independent given $A$, with $A$ and $B$ binary and $Y$ continuous. The conditional independence assumption is represented by the directed acyclic graph structure in Figure 1(a). Let $\theta_a = P(A = a)$, $\theta_{b|a} = P(B = b \,|\, A = a)$, and let $p(y \,|\, \beta_a, \sigma_a)$ be the conditional density of $Y$ given $A = a$, assumed normal with mean $\beta_a$ and variance $\sigma_a^2$. We want to estimate the probability that condition $A$ is present given specified results from the two tests $B$ and $Y$. Let $\boldsymbol{\Theta}$ represent all of the parameters. If $\boldsymbol{\Theta}$ were known, we would use the formula:

$$q(\boldsymbol{\Theta}) = q_{a|b,y}(\boldsymbol{\Theta}) = \frac{\theta_a \theta_{b|a} p(y \,|\, \beta_a, \sigma_a)}{\sum_{a_1} \theta_{a_1} \theta_{b|a_1} p(y \,|\, \beta_{a_1}, \sigma_{a_1})} \,. \quad (1)$$

In the Bayesian paradigm, uncertainty about $\boldsymbol{\Theta}$ is quantified by modeling parameters as random variables. It follows that query probabilities such as (1) are also random. A query response is usually estimated by approximating its posterior mean. This approximation is similar to expression (1), but with $\theta_a$ and $\theta_{b|a}$ replaced by their posterior means and with the normal densities replaced by Student's $t$ densities.

One may want more than just a point estimate. Van Allen *et al.* (2001, 2008) showed (for discrete networks) how one can approximate the variance and posterior distribution of a query. Their variance derivation employs the delta method; i.e., a first-order Taylor series expansion of the function $q(\boldsymbol{\Theta})$ about the posterior mean of $\boldsymbol{\Theta}$. They provide asymptotic theory and empirical experiments supporting this approach. They also showed how these approximations can be used to construct a Bayesian credible interval (error bars) for $q(\boldsymbol{\Theta})$. Guo and Greiner (2005) applied this delta method approximation as part of a mean squared error (i.e., squared bias + variance) measure designed to estimate the quality of different belief net structures when seeking a best classifier. Lee *et al.* (2006) provide a technique for combining independent belief net classifiers that involves weighting their respective mean probability values by their inverse variances, and they show that this works well in practice.

We propose new approximations for the mean and variance based on a simple trick. Suppose $(A_1, B_1, Y_1)$ and $(A_2, B_2, Y_2)$ are replicates of the network variables, conditionally independent given $\boldsymbol{\Theta}$. We represent the paired replicates as nodes in a "doubled network" with the same structure; see Figure 1. The squared query $q(\boldsymbol{\Theta})^2$ can be expressed as a query in this doubled net-

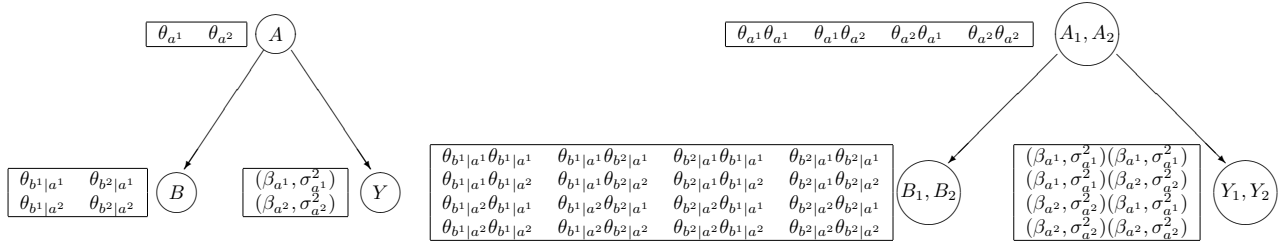

Figure 1: (a) A simple Bayesian network. (b) The corresponding doubled network.

work:
$$P(A_1 = A_2 = a \mid B_1 = B_2 = b, Y_1 = Y_2 = y, \boldsymbol{\Theta}).$$

The method used to approximate the mean of $q(\boldsymbol{\Theta})$ can be extended to the doubled network to approximate the mean of $q(\boldsymbol{\Theta})^2$ and hence to approximate the variance. Unlike the delta method, our approach does not rely on approximate local linearity of $q(\boldsymbol{\Theta})$. It does involve the addition of two incomplete observations to the data set when calculating the posterior mean of $q(\boldsymbol{\Theta})^2$. In some situations, this addition results in under-estimation of the desired variance. This deficiency is largely eliminated by a simple adjustment. A similar adjustment substantially improves the usual query mean approximation.

Section 2 reviews pertinent models and methods for belief networks. The network doubling technique is described in Section 3 for discrete, continuous, and hybrid networks. Proposed adjustments and numerical results are presented in Sections 4 and 5 for discrete networks. Corresponding work for continuous and hybrid networks is ongoing. Computational issues are discussed in Section 6. Contributions and plans for further work are summarized in Section 7.

## 2 BACKGROUND

### 2.1 NETWORK VARIABLES

We assume network structure is known. Let $B$ denote a discrete network variable taking values $b \in \text{Dom}_B$. Let $Y$ denote a continuous network variable taking values $y$ on the real line. Vectors of variables are denoted by boldface: $\boldsymbol{A}$ for discrete and $\boldsymbol{X}$ for continuous. Let $\boldsymbol{\Theta}$ be a random vector comprising all unknown network parameters; i.e., $\boldsymbol{\Theta}$ determines all conditional distributions of variables given their parents.

We assume that discrete variables have only discrete parents. Suppose $pa(B) = \boldsymbol{A}$; i.e., the parents of $B$ are the variables comprising the vector $\boldsymbol{A}$. The conditional probability that $B = b$ given $\boldsymbol{A} = \boldsymbol{a}$ is denoted
$$\theta_{b|\boldsymbol{a}} = \theta_{B=b|\boldsymbol{A}=\boldsymbol{a}} = P\{B = b \mid \boldsymbol{A} = \boldsymbol{a}, \boldsymbol{\Theta}\}.$$

Variables associated with values will be clear from context. We employ similar abbreviations for other parameters and hyperparameters. The $\theta_{b|\boldsymbol{a}}$ parameters are often presented in conditional probability tables (CPtables) with rows indexed by $\boldsymbol{a}$ and columns by $b$; e.g., see Figure 1. Note that we use superscripts $b^1, b^2$ to list the distinct values in $\text{Dom}_B$. We use subscripts $b_1, b_2$ to denote arbitrary values in $\text{Dom}_B$, often related to replicated variables $B_1, B_2$.

Continuous variables can have both discrete and continuous parents. Suppose $pa(Y) = \langle \boldsymbol{A}, \boldsymbol{X} \rangle$ with $\boldsymbol{X} = \langle X_1, \ldots, X_d \rangle$. The conditional distribution of $Y$ is
$$(Y \mid \boldsymbol{A} = \boldsymbol{a}, \boldsymbol{X} = \boldsymbol{x}, \boldsymbol{\Theta}) \sim N\left((1, \boldsymbol{x}^T)\boldsymbol{\beta_a}, \sigma_{\boldsymbol{a}}^2\right); \quad (2)$$
i.e., normally distributed, conditional mean related to $\boldsymbol{x}$ by a linear regression model with coefficients depending on $\boldsymbol{a}$. Here $\boldsymbol{x}^T$ is the transpose of the $d$-dimensional column vector $\boldsymbol{x}$ while $\boldsymbol{\beta_a}$ is an $(d+1)$-dimensional column vector of regression coefficients (the first entry is the constant term).

### 2.2 PRIOR AND POSTERIOR

The network parameters represented by $\boldsymbol{\Theta}$ consist of CPtable parameters $\theta_{b|\boldsymbol{a}}$, regression coefficient vectors $\boldsymbol{\beta_a}$, and variances $\sigma_{\boldsymbol{a}}^2$. We assume the prior distribution for $\boldsymbol{\Theta}$ has the following form; e.g., see Gelman *et al.* (2003).

- CPtable rows follow Dirichlet distributions:
$$\boldsymbol{\theta}_{B|\boldsymbol{a}} := \langle \theta_{b|\boldsymbol{a}}, b \in \text{Dom}_B \rangle \sim \text{Dir}(\boldsymbol{\alpha}_{B|\boldsymbol{a}}),$$
where $\boldsymbol{\alpha}_{B|\boldsymbol{a}} := \langle \alpha_{b|\boldsymbol{a}}, b \in \text{Dom}_B \rangle$.

- The regression coefficients and variance together have a normal-(inverse chi-square) distribution:
$$\begin{aligned} (\boldsymbol{\beta_a} \mid \sigma_{\boldsymbol{a}}^2) &\sim N_{d+1}\left(\boldsymbol{\mu_a}, \sigma_{\boldsymbol{a}}^2(\nu_{\boldsymbol{a}} \boldsymbol{\Psi_a})^{-1}\right), \\ \sigma_{\boldsymbol{a}}^{-2} &\sim (\tau_{\boldsymbol{a}}^2 \nu_{\boldsymbol{a}})^{-1} \chi_{\nu_{\boldsymbol{a}}}^2. \end{aligned}$$

I.e., dropping subscripts for a moment, $\boldsymbol{\beta}$ conditioned on $\sigma^2$ is multivariate normal with mean vector $\boldsymbol{\mu}$ and covariance matrix $\sigma^2(\nu\boldsymbol{\Psi})^{-1}$; and $\nu\tau^2/\sigma^2$ has a $\chi^2_\nu$ distribution with $\nu > 0$ (not necessarily an integer). Note that $\tau^2/\sigma^2$ has mean 1 and variance $2/\nu$.

- Parameters are assumed to be statistically independent except where joint distributions are specified above. In particular, we assume global independence: the parameters determining the conditional distribution of one variable given its parents are independent of all other parameters.

The prior is conjugate: given a data set $\mathcal{D}$ consisting of $n$ independent replicates of complete tuples of network variables, the prior hyperparameter values are updated as follows. Let $n_{ab}$ and $n_a$ be the number of tuples in $\mathcal{D}$ with $(\boldsymbol{A}, B) = (\boldsymbol{a}, b)$ and $\boldsymbol{A} = \boldsymbol{a}$, respectively. Let $(\boldsymbol{x}_i, y_i)$ be the observations of $(\boldsymbol{X}, Y)$ for the $n_{\boldsymbol{a}}$ tuples with $\boldsymbol{A} = \boldsymbol{a}$. Let $\boldsymbol{X}_{\boldsymbol{a}}$ be the $n_{\boldsymbol{a}} \times (d+1)$ matrix with rows $(1, \boldsymbol{x}_i^T)$. Let $\boldsymbol{y}_{\boldsymbol{a}}$ be the column vector with entries $y_i$. In the five equations below, the prior hyperparameter values appear on the right-hand side and are identified with tildes (e.g., $\tilde{\alpha}$).

$$\begin{aligned}
\alpha_{b|\boldsymbol{a}} &= \tilde{\alpha}_{b|\boldsymbol{a}} + n_{ab} \\
\nu_{\boldsymbol{a}} &= \tilde{\nu}_{\boldsymbol{a}} + n_{\boldsymbol{a}} \\
\nu_{\boldsymbol{a}}\boldsymbol{\Psi}_{\boldsymbol{a}} &= \tilde{\nu}_{\boldsymbol{a}}\tilde{\boldsymbol{\Psi}}_{\boldsymbol{a}} + \boldsymbol{X}_{\boldsymbol{a}}^T\boldsymbol{X}_{\boldsymbol{a}} \\
\nu_{\boldsymbol{a}}\boldsymbol{\Psi}_{\boldsymbol{a}}\boldsymbol{\mu}_{\boldsymbol{a}} &= \tilde{\nu}_{\boldsymbol{a}}\tilde{\boldsymbol{\Psi}}_{\boldsymbol{a}}\tilde{\boldsymbol{\mu}}_{\boldsymbol{a}} + \boldsymbol{X}_{\boldsymbol{a}}^T\boldsymbol{y}_{\boldsymbol{a}} \\
\nu_{\boldsymbol{a}}\left[\tau_{\boldsymbol{a}}^2 + \boldsymbol{\mu}_{\boldsymbol{a}}^T\boldsymbol{\Psi}_{\boldsymbol{a}}\boldsymbol{\mu}_{\boldsymbol{a}}\right] &= \tilde{\nu}_{\boldsymbol{a}}\left[\tilde{\tau}_{\boldsymbol{a}}^2 + \tilde{\boldsymbol{\mu}}_{\boldsymbol{a}}^T\tilde{\boldsymbol{\Psi}}_{\boldsymbol{a}}\tilde{\boldsymbol{\mu}}_{\boldsymbol{a}}\right] + \boldsymbol{y}_{\boldsymbol{a}}^T\boldsymbol{y}_{\boldsymbol{a}}
\end{aligned}$$

The values $\sum_{\boldsymbol{a},b} \alpha_{b|\boldsymbol{a}}$ and $\sum_{\boldsymbol{a}} \nu_{\boldsymbol{a}}$ are called the effective sample sizes for variables $B$ and $Y$, respectively. Our adjustments developed in Section 4 are motivated by large $m$ asymptotics, where $m$ is proportional to the effective sample size for each of the variables; i.e.,

$$\alpha_{b|\boldsymbol{a}} = m\alpha_{b|\boldsymbol{a}}^0 \text{ and } \nu_{\boldsymbol{a}} = m\nu_{\boldsymbol{a}}^0$$
$$\text{with } (\alpha_{b|\boldsymbol{a}}^0, \nu_{\boldsymbol{a}}^0, \boldsymbol{\Psi}_{\boldsymbol{a}}, \boldsymbol{\mu}_{\boldsymbol{a}}, \tau_{\boldsymbol{a}}^2) \text{ fixed.} \quad (3)$$

Large $m$ asymptotics are similar to but not the same as large $n$ asymptotics. As the sample size $n$ increases, the posterior mean $E\{\theta_{b|\boldsymbol{a}} \mid \mathcal{D}\} = \alpha_{b|\boldsymbol{a}}/\alpha_{\cdot|\boldsymbol{a}}$ varies and converges to some value. (Here and elsewhere, the dot subscript indicates summation: $\alpha_{\cdot|\boldsymbol{a}} = \sum_b \alpha_{b|\boldsymbol{a}}$.) Under assumption (3), the posterior mean remains fixed as $m$ varies.

### 2.3 APPROXIMATING A QUERY MEAN

Consider a query involving outcomes of hypothesis variables $\boldsymbol{H}$ given values for evidence variables $\boldsymbol{E}$. It is convenient to represent the query in terms of a function $w(\boldsymbol{H})$. E.g., suppose $\boldsymbol{H} = A$, $\boldsymbol{E} = (B, Y)$, $\boldsymbol{e} = (b, y)$, and

$$\begin{aligned}
q(\boldsymbol{\Theta}) &= P(A = a \mid B = b, Y = y, \boldsymbol{\Theta}) \\
&= E\{w(A) \mid B = b, Y = y, \boldsymbol{\Theta}\},
\end{aligned}$$

where $w(A) = 1$ for $A = a$ and $w(A) = 0$ otherwise.

For discrete networks, query responses $q(\boldsymbol{\Theta})$ are usually estimated by $q(\hat{\boldsymbol{\Theta}})$, where $\hat{\boldsymbol{\Theta}} := E\{\boldsymbol{\Theta} \mid \mathcal{D}\}$ is the posterior mean of the parameter vector. This plug-in estimate usually differs slightly from the posterior query mean $E\{q(\boldsymbol{\Theta}) \mid \mathcal{D}\}$. Cooper and Herskovits (1992, expression 19) showed that the plug-in estimate equals $E\{q(\boldsymbol{\Theta}) \mid \mathcal{D}, \boldsymbol{e}\}$; i.e., the posterior query mean given an augmented data set consisting of $\mathcal{D}$ and an additional partial observation of the evidence variables $\boldsymbol{E} = \boldsymbol{e}$. Cooper and Herskovits (1991) derived a formula for $E\{q(\boldsymbol{\Theta}) \mid \mathcal{D}, \boldsymbol{e}\}$ that is valid for discrete, continuous, and hybrid networks. This formula provides a useful approximation of the less tractable $E\{q(\boldsymbol{\Theta}) \mid \mathcal{D}\}$. The plug-in estimate is a special case of this formula for discrete networks. The formula is important for our network doubling technique, so is reviewed here.

In the integral expression below, $\boldsymbol{Z}$ represents all variables not included in $(\boldsymbol{H}, \boldsymbol{E})$; $d\boldsymbol{h}$ and $d\boldsymbol{z}$ refer to product measures allowing both integration for continuous variables (Lebesgue measure) and summation for discrete variables (counting measure). Some manipulation yields

$$\begin{aligned}
E\{q(\boldsymbol{\Theta}) \mid \mathcal{D}, \boldsymbol{e}\} &= E\{w(\boldsymbol{H}) \mid \boldsymbol{E} = \boldsymbol{e}, \mathcal{D}\} \\
&= E\left[E\{w(\boldsymbol{H}) \mid \boldsymbol{E} = \boldsymbol{e}, \boldsymbol{\Theta}\} \mid \mathcal{D}\right] \quad (4) \\
&= \frac{\int\int w(\boldsymbol{h}) \int p(\boldsymbol{h},\boldsymbol{e},\boldsymbol{z} \mid \boldsymbol{\theta})p(\boldsymbol{\theta} \mid \mathcal{D})d\boldsymbol{\theta}d\boldsymbol{h}d\boldsymbol{z}}{\int\int\int p(\boldsymbol{h},\boldsymbol{e},\boldsymbol{z} \mid \boldsymbol{\theta})p(\boldsymbol{\theta} \mid \mathcal{D})d\boldsymbol{\theta}d\boldsymbol{h}d\boldsymbol{z}}.
\end{aligned}$$

Now $p(\boldsymbol{h}, \boldsymbol{e}, \boldsymbol{z} \mid \boldsymbol{\theta})$ factors as a product of conditional probabilities and densities, one for each variable in the network. Due to global independence, the integral $\int p(\boldsymbol{h}, \boldsymbol{e}, \boldsymbol{z} \mid \boldsymbol{\theta})p(\boldsymbol{\theta} \mid \mathcal{D})d\boldsymbol{\theta}$ factors into a product of integrals, one for each variable. The result is a product of probabilities and densities described in Section 2.4 below. It follows that $E\{q(\boldsymbol{\Theta}) \mid \mathcal{D}, \boldsymbol{e}\}$ can be calculated in essentially the same manner as the function $q(\boldsymbol{\Theta})$, but with two modifications.

- For discrete variables, parameters $\theta_{b|\boldsymbol{a}}$ are replaced by their posterior means. If all network variables are discrete, then we have the plug-in estimate:

$$E\{q(\boldsymbol{\Theta}) \mid \mathcal{D}, \boldsymbol{e}\} = q(E\{\boldsymbol{\Theta} \mid \mathcal{D}\}). \quad (5)$$

- For continuous variables, the normal densities are replaced by the $St_1(\eta, \omega^2, \nu)$ densities described below. Note that this is *not* the same as replacing $\boldsymbol{\beta}$ and $\sigma^2$ parameters with their posterior means.

### 2.4 PREDICTIVE DISTRIBUTIONS

The predictive distribution of the network variables is obtained by integrating out their joint conditional dis-tribution given $\boldsymbol{\Theta}$ with respect to the posterior distribution of $\boldsymbol{\Theta}$. Global independence allows this integration to be carried out separately for each conditional distribution of a variable given its parents.

The predictive distribution for a discrete variable $B$ is

$$\pi_{b|\boldsymbol{a}} := P(B=b \,|\, \boldsymbol{A}=\boldsymbol{a}, \mathcal{D}) = E\{\theta_{b|\boldsymbol{a}} \,|\, \mathcal{D}\} = \frac{\alpha_{b|\boldsymbol{a}}}{\alpha_{\cdot|\boldsymbol{a}}}.$$

The predictive distribution for a continuous variable is a location-scale version of the Student's $t$ distribution with $\nu$ degrees of freedom. We need the multivariate form of this distribution in Section 3, so we define it here. Suppose

$$\boldsymbol{T} = \boldsymbol{\eta} + U^{-1/2}(\boldsymbol{Z} - \boldsymbol{\eta}),$$

where $\boldsymbol{Z}$ and $U$ are independent, $\boldsymbol{Z} \sim N_p(\boldsymbol{\eta}, \boldsymbol{\Omega})$, $U \sim (1/\nu)\chi^2_\nu$, and $\boldsymbol{\Omega}$ is a nonsingular covariance matrix. It follows that $\boldsymbol{T}$ has the following density function (Johnson and Kotz, 1972, page 134):

$$\frac{\Gamma[(\nu+p)/2]\,/\,\Gamma(\nu/2)}{(\nu\pi)^{p/2}|\boldsymbol{\Omega}|^{1/2}\left[1 + \frac{1}{\nu}(\boldsymbol{t}-\boldsymbol{\eta})^T\boldsymbol{\Omega}^{-1}(\boldsymbol{t}-\boldsymbol{\eta})\right]^{(\nu+p)/2}}.$$

We refer to this as the $St_p(\boldsymbol{\eta}, \boldsymbol{\Omega}, \nu)$ distribution. For $p=1$, we write $St_1(\eta, \omega^2, \nu)$. Note that $St_1(0,1,\nu)$ is Student's $t$ distribution.

We claim that $(Y \,|\, \boldsymbol{A}=\boldsymbol{a}, \boldsymbol{X}=\boldsymbol{x}, \mathcal{D}) \sim St_1(\eta, \omega^2, \nu)$ with $\nu = \nu_{\boldsymbol{a}}$, $\eta = (1, \boldsymbol{x}^T)\boldsymbol{\mu}_{\boldsymbol{a}}$, and

$$\omega^2 = \tau_{\boldsymbol{a}}^2 \left\{(1, \boldsymbol{x}^T)(\nu_{\boldsymbol{a}}\boldsymbol{\Psi}_{\boldsymbol{a}})^{-1}(1, \boldsymbol{x}^T)^T + 1\right\}. \quad (6)$$

To see this, let us suppress subscripts for a moment. Let $Z_1 \sim N(0,1)$ be independent of $(\boldsymbol{\beta}, \sigma)$. Put $\boldsymbol{Z}_2 := \sigma^{-1}(\boldsymbol{\beta} - \boldsymbol{\mu}) \sim N_{m+1}\left(\boldsymbol{0}, (\nu\boldsymbol{\Psi})^{-1}\right)$. We then have

$$\begin{aligned}(Y \,|\, \boldsymbol{a}, \boldsymbol{x}, \mathcal{D}) &\sim (1, \boldsymbol{x}^T)\boldsymbol{\beta} + \sigma Z_1 \\ &\sim \eta + (\sigma/\tau)\tau\left\{(1, \boldsymbol{x}^T)\boldsymbol{Z}_2 + Z_1\right\}.\end{aligned}$$

## 3  NETWORK DOUBLING

In Section 2.3 we noted that $E\{q(\boldsymbol{\Theta}) \,|\, \mathcal{D}\}$ is usually approximated by the more tractable $E\{q(\boldsymbol{\Theta}) \,|\, \mathcal{D}, \boldsymbol{e}\}$. Here we propose approximating $\mathrm{Var}\{q(\boldsymbol{\Theta}) \,|\, \mathcal{D}\}$ by $\mathrm{Var}\{q(\boldsymbol{\Theta}) \,|\, \mathcal{D}, \boldsymbol{e}, \boldsymbol{e}\}$; i.e., the posterior variance given $\mathcal{D}$ and additional replicates $\boldsymbol{E}_1$ and $\boldsymbol{E}_2$ of the vector of evidence variables, both having the same value $\boldsymbol{e}$. We develop a formula for this latter variance by imagining a doubled network; see Figure 1(b). These mean and variance approximations can be improved by adjustments described in Section 4.

Consider two replicated tuples of network variables, conditionally independent and identically distributed given $\boldsymbol{\Theta}$. Use these to replace each variable in the original network by a pair of variables; e.g., $B$ is replaced by $B^* := (B_1, B_2)$ with possible values $b^* = (b_1, b_2) \in \mathrm{Dom}_{B^*} = \mathrm{Dom}_B \times \mathrm{Dom}_B$. If $pa(B) = \boldsymbol{A}$, then $pa(B^*) = \boldsymbol{A}^* := (\boldsymbol{A}_1, \boldsymbol{A}_2)$. Conditional distributions of doubled variables given parents are obtained by multiplying probabilities or densities for single variables.

For discrete variables, we have

$$P(B^* = b^* \,|\, \boldsymbol{A}^* = \boldsymbol{a}^*, \boldsymbol{\Theta}) = \theta_{b_1|\boldsymbol{a}_1}\theta_{b_2|\boldsymbol{a}_2}.$$

E.g., if $\boldsymbol{A} = A$, $\mathrm{Dom}_A = \{a^1, a^2\}$, and $\mathrm{Dom}_B = \{b^1, b^2\}$, then the CPtable for $B^*$ is the $4 \times 4$ array shown in Figure 1(b). More generally, if a CPtable in the original network involves $d_r \times d_c$ parameters, then corresponding table in the doubled network has $d_r^2 \times d_c^2$ entries. Note that CPtable rows in the doubled network are not independent (local independence does not hold) and do not have Dirichlet distributions. Fortunately, these properties are not needed for the factorization described following (4).

For continuous variables, the conditional density of $Y^* = (Y_1, Y_2)$ given $(\boldsymbol{A}^* = \boldsymbol{a}^*, \boldsymbol{X}^* = \boldsymbol{x}^*, \boldsymbol{\Theta})$ is the product of the densities for two normal distributions of the form (2) with subscript $i = 1, 2$ on $\boldsymbol{a}$ and $\boldsymbol{x}$.

Put $\boldsymbol{H}^* = (\boldsymbol{H}_1, \boldsymbol{H}_2)$, $w^*(\boldsymbol{H}^*) = w(\boldsymbol{H}_1)w(\boldsymbol{H}_2)$, $\boldsymbol{E}^* = (\boldsymbol{E}_1, \boldsymbol{E}_2)$, and $\boldsymbol{e}^* = (\boldsymbol{e}, \boldsymbol{e})$. Some manipulation using conditional independence yields

$$\begin{aligned}q(\boldsymbol{\Theta})^2 &= E\{w^*(\boldsymbol{H}^*) \,|\, \boldsymbol{E}^* = \boldsymbol{e}^*, \boldsymbol{\Theta}\}, \\ q(\boldsymbol{\Theta}) &= E\{w(H_1) \,|\, \boldsymbol{E}^* = \boldsymbol{e}^*, \boldsymbol{\Theta}\}.\end{aligned}$$

We thus have

$$\begin{aligned}\mathrm{Var}&\{q(\boldsymbol{\Theta}) \,|\, \mathcal{D}, \boldsymbol{e}, \boldsymbol{e}\} \quad (7) \\ &= E\{q(\boldsymbol{\Theta})^2 \,|\, \mathcal{D}, \boldsymbol{e}, \boldsymbol{e}\} - [E\{q(\boldsymbol{\Theta}) \,|\, \mathcal{D}, \boldsymbol{e}, \boldsymbol{e}\}]^2 \\ &= E\{w^*(\boldsymbol{H}^*) \,|\, \boldsymbol{e}^*, \mathcal{D}\} - [E\{w(H_1) \,|\, \boldsymbol{e}^*, \mathcal{D}\}]^2.\end{aligned}$$

The doubled network satisfies global independence assumptions, so we can follow the approach of Section 2.3 to evaluate the two expected values in (7). To accomplish this task, we need bivariate predictive distributions for the doubled network.

For discrete variables, the calculation follows from the means and covariances of a Dirichlet distribution. Let $\delta_{b_1 b_2}$ be the Kronecker delta function. We have

$$\begin{aligned}\pi^*_{b^*|\boldsymbol{a}^*} &:= P\{B^* = b^* \,|\, \boldsymbol{A}^* = \boldsymbol{a}^*, \mathcal{D}\} \\ &= E\{\theta_{b_1|\boldsymbol{a}_1}\theta_{b_2|\boldsymbol{a}_2} \,|\, \mathcal{D}\} \\ &= \pi_{b_1|\boldsymbol{a}_1}\pi_{b_2|\boldsymbol{a}_2} + \delta_{\boldsymbol{a}_1\boldsymbol{a}_2}\frac{\pi_{b_1|\boldsymbol{a}_1}(\delta_{b_1 b_2} - \pi_{b_2|\boldsymbol{a}_1})}{\alpha_{\cdot|\boldsymbol{a}_1} + 1}.\end{aligned}$$

If all network variables are discrete, then we have an identity corresponding to (5). Let $\boldsymbol{\Theta}^*$ be the vector of all CPtable entries in the doubled network; e.g., $\theta_{b_1|a_1}\theta_{b_2|a_2}$ appears in row $a^*$ and column $b^*$ for the CPtable of $B^*$. We then have

$$E\{q^*(\boldsymbol{\Theta}^*) \,|\, \mathcal{D}, \boldsymbol{e}, \boldsymbol{e}\} = q^*(E\{\boldsymbol{\Theta}^* \,|\, \mathcal{D}\}) \qquad (8)$$

with the entries in $E\{\boldsymbol{\Theta}^* \,|\, \mathcal{D}\}$ given by the $\pi^*_{b^*|a^*}$ values above. The two expected values in the variance approximation (7) are calculated by applying (8) twice: with $q^*(\boldsymbol{\Theta}^*) = q(\boldsymbol{\Theta})^2$ and with $q^*(\boldsymbol{\Theta}^*) = q(\boldsymbol{\Theta})$.

For continuous variables, we need the density for $\{(Y_1, Y_2) \,|\, \boldsymbol{a}_1, \boldsymbol{a}_2, \boldsymbol{x}_1, \boldsymbol{x}_2, \mathcal{D}\}$. There are two cases to consider.

- If $\boldsymbol{a}_1 \neq \boldsymbol{a}_2$, then the parameters $(\boldsymbol{\beta}_{\boldsymbol{a}_1}, \sigma^2_{\boldsymbol{a}_1})$ and $(\boldsymbol{\beta}_{\boldsymbol{a}_2}, \sigma^2_{\boldsymbol{a}_2})$ are mutually independent. Consequently, the joint distribution factors as a product of two $St_1(\eta, \omega^2, \nu)$ densities; see expression (6).

- If $\boldsymbol{a}_1 = \boldsymbol{a}_2$ ($= \boldsymbol{a}$, say), then the joint distribution is $St_2(\boldsymbol{\eta}, \boldsymbol{\Omega}, \nu)$ with $\nu = \nu_{\boldsymbol{a}}$, $\boldsymbol{\eta} = \boldsymbol{X}_2 \boldsymbol{\mu}_{\boldsymbol{a}}$, and

$$\boldsymbol{\Omega} = \tau^2_{\boldsymbol{a}} \left\{ \boldsymbol{X}_2 (\nu_{\boldsymbol{a}} \boldsymbol{\Psi}_{\boldsymbol{a}})^{-1} \boldsymbol{X}_2^T + \boldsymbol{I}_2 \right\},$$

where $\boldsymbol{X}_2$ is the $2 \times (1+d)$ matrix whose rows are each $(1, \boldsymbol{x}_i^T)$ and $\boldsymbol{I}_2$ is the $2 \times 2$ identity matrix. The derivation is similar to that following (6). Note that $(\boldsymbol{\beta}_{\boldsymbol{a}}, \sigma^2_{\boldsymbol{a}})$ is the same for both $Y_1$ and $Y_2$ in this case.

## 4 ADJUSTMENTS

We now narrow our focus to discrete networks and consider the four mean and variance approximations in Table 1. The delta method approximation is

$$\hat{v}_1 = \boldsymbol{g}^T \boldsymbol{C} \boldsymbol{g}, \qquad (9)$$

where $\boldsymbol{g}$ is the gradient vector of $q(\boldsymbol{\Theta})$ and $\boldsymbol{C}$ is the covariance matrix of $\boldsymbol{\Theta}$, both evaluated at $E\{\boldsymbol{\Theta} \,|\, \mathcal{D}\}$. The second variance approximation $\hat{v}_2$ is the doubling method introduced in Section 3. The simple adjustments $(\hat{q}_3, \hat{v}_3)$ and more complex adjustments $(\hat{q}_4, \hat{v}_4)$ are developed in this section.

For conciseness we suppress $\mathcal{D}$ in our expressions; i.e., we implicitly assume that expectations are conditioned on $\mathcal{D}$. Put $Q = q(\boldsymbol{\Theta}) = P(\boldsymbol{H} = \boldsymbol{h} \,|\, \boldsymbol{E} = \boldsymbol{e}, \boldsymbol{\Theta})$ and $R = P(\boldsymbol{E} = \boldsymbol{e} \,|\, \boldsymbol{\Theta})$. Note that $R$ is an unconditional query, with hypothesis $\boldsymbol{E} = \boldsymbol{e}$ and no evidence variables. Let $\mu_q$, $\mu_r$, $\sigma_{qq}$, $\sigma_{rr}$, and $\sigma_{qr}$ denote the means, variances, and covariance for $(Q, R)$. We extend this notation to higher moments; e.g., $\sigma_{qqr} = E\{(Q-\mu_q)^2(R-\mu_r)\}$.

We use approximations for higher moments motivated by large $m$ asymptotics; i.e., a sequence of posterior distributions of the form (3) with $m \to \infty$. One may verify that the distribution of $\sqrt{m}(Q - \mu_q, R - \mu_r)$ converges to bivariate normal by modifying the proof of Theorem 2 in Van Allen et al. (2008). Asymptotic normality implies that

$$\sigma_{qqrr} - 2\sigma^2_{qr} - \sigma_{qq}\sigma_{rr} \to 0 \text{ at rate } m^{-5/2} \qquad (10)$$

while $\sigma_{qrr}$ and $\sigma_{qqr}$ converge to zero at rate $m^{-2}$. We considered approximating $\sigma_{qqr}$ and $\sigma_{qrr}$ by zero but found that more accurate approximations give better results. Asymptotic bivariate normality suggests

$$E\{R - \mu_r \,|\, Q\} \approx (Q - \mu_q)\frac{\sigma_{qr}}{\sigma_{qq}}$$

and hence $\sigma_{qqr} \approx \sigma_{qqq}\sigma_{qr}/\sigma_{qq}$. Now $\sigma_{qqq} = 0$ for normal distributions; however, Van Allen et al. (2008) argue that query distributions are usually better approximated by beta distributions. Substituting the third moment of a beta distribution for $\sigma_{qqq}$, we obtain

$$\sigma_{qqr} \approx \frac{2\sigma_{qr}\sigma_{qq}(1 - 2\mu_q)}{\mu_q(1 - \mu_q) + \sigma_{qq}}. \qquad (11)$$

Table 1: Summary of approximations for $\mu_q$ and $\sigma_{qq}$.

| Means | Variances |
| --- | --- |
| $\hat{q}_1 = E\{q(\boldsymbol{\Theta}) \,|\, \mathcal{D}, \boldsymbol{e}\}$ | $\hat{v}_1 =$ delta method (9) |
| $\hat{q}_2 = E\{q(\boldsymbol{\Theta}) \,|\, \mathcal{D}, \boldsymbol{e}, \boldsymbol{e}\}$ | $\hat{v}_2 = \mathrm{Var}\{q(\boldsymbol{\Theta}) \,|\, \mathcal{D}, \boldsymbol{e}, \boldsymbol{e}\}$ |
| $\hat{q}_3 = \hat{q}_1 - (\hat{q}_2 - \hat{q}_1)$ | $\hat{v}_3 =$ expression (18) |
| $\hat{q}_4 = \hat{q}_1 - \hat{\sigma}_{qr}/\mu_r$ | $\hat{v}_4 =$ expression (17) |

Switching the roles of $Q$ and $R$ gives

$$\sigma_{qrr} \approx \frac{2\sigma_{qr}\sigma_{rr}(1 - 2\mu_r)}{\mu_r(1 - \mu_r) + \sigma_{rr}}. \qquad (12)$$

Before proceeding, we observe that $\mu_r$ and $\sigma_{rr}$ can be calculated exactly because $R$ can be expressed as a sum of products of independent terms. For queries with this property, all approximations (except $\hat{v}_1$) are exact; i.e., additional observations of evidence variables have no effect on the posterior mean or variance. E.g., given a discrete network with structure $E \to B \to H$, we have $q(\boldsymbol{\Theta}) = \sum_b \theta_{h|b}\theta_{b|e}$. Since parameters in each product are independent, it follows that $\hat{q}_2 = \hat{q}_1 = \mu_q$ and $\hat{v}_2 = \sigma_{qq}$.

We begin with adjustments to improve $\hat{q}_1$. Bayes rule and some manipulation yields

$$\hat{q}_1 = \frac{E(QR)}{E(R)} = \mu_q + \frac{\sigma_{qr}}{\mu_r} \qquad (13)$$

$$\hat{q}_2 = \frac{E(QR^2)}{E(R^2)} = \mu_q + \frac{2\mu_r\sigma_{qr} + \sigma_{qrr}}{\mu_r^2 + \sigma_{rr}}.$$

If $\mu_r = 1$, then set $\hat{\sigma}_{qr} = 0$. Otherwise, substituting (12) for $\sigma_{qrr}$ and solving yields $\hat{\sigma}_{qr} =$

$$\frac{(\hat{q}_2 - \hat{q}_1)\mu_r(\mu_r^2 + \sigma_{rr})\{\mu_r(1 - \mu_r) + \sigma_{rr}\}}{\mu_r^3(1 - \mu_r) + \mu_r(1 - 2\mu_r)\sigma_{rr} - \sigma_{rr}^2}. \quad (14)$$

The formula for $\hat{q}_4$ in Table 1 follows from (13). Now recall that, under condition (3), $\mu_r$ remains fixed while $\sigma_{rr} \to 0$ as $m \to \infty$. It follows that setting $\sigma_{rr} = 0$ in (14) will have negligible effect for large $m$. We thus obtain $\hat{\sigma}_{qr} \approx (\hat{q}_2 - \hat{q}_1)\mu_r$, leading to the simpler $\hat{q}_3$ approximation.

In trying to improve $\hat{v}_2$, we began with the idea of replacing $\hat{q}_2$ with $\mu_q$:

$$E\{(Q - \mu_q)^2 \,|\, \boldsymbol{e}, \boldsymbol{e}\} = \hat{v}_2 + (\hat{q}_2 - \mu_q)^2. \quad (15)$$

This suggests an approximation $\hat{v}_2 + 4(\hat{q}_2 - \hat{q}_1)^2$, which does help to reduce the under-estimation problem; however, a greater improvement is obtained by further analysis of (15):

$$\frac{E\{(Q - \mu_q)^2 R^2\}}{E(R^2)} = \frac{\mu_r^2\sigma_{qq} + 2\mu_r\sigma_{qqr} + \sigma_{qqrr}}{\mu_r^2 + \sigma_{rr}}. \quad (16)$$

We approximate $\sigma_{qqrr}$ using (10), $\sigma_{qqr}$ by (11), $\sigma_{qr}$ by (14), $\mu_q$ by $\hat{q}_4$, and replace $\sigma_{qq}$ by $\hat{v}_4$. Rearranging terms yields the identity: $\hat{v}_4 =$

$$\frac{(\mu_r^2 + \sigma_{rr})\{\hat{v}_2 + (\hat{q}_2 - \hat{q}_4)^2\} - 2\hat{\sigma}_{qr}^2}{\mu_r^2 + \sigma_{rr} + 4\mu_r\hat{\sigma}_{qr}(1 - 2\hat{q}_4)/\{\hat{q}_4(1 - \hat{q}_4) + \hat{v}_4\}}. \quad (17)$$

Notice that $\hat{v}_4$ appears in the denominator of (17). We initially set this value to $\hat{v}_2$, then iteratively solve for $\hat{v}_4$. The values converge in a few iterations.

We observe that replacing $\sigma_{rr}$ by zero has negligible effect on (17) as $m \to \infty$. By also replacing $\hat{q}_4$ by $\hat{q}_3$ and $\hat{\sigma}_{qr}/\mu_r$ by $\hat{q}_2 - \hat{q}_1$, we obtain a simpler identity:

$$\hat{v}_3 = \frac{\hat{v}_2 + 2(\hat{q}_2 - \hat{q}_1)^2}{1 + 4(\hat{q}_2 - \hat{q}_1)(1 - 2\hat{q}_3)/\{\hat{q}_3(1 - \hat{q}_3) + \hat{v}_3\}}. \quad (18)$$

We again initialize by $\hat{v}_2$, then iteratively solve for $\hat{v}_3$. The approximations $\hat{q}_3$ and $\hat{v}_3$ may be preferred to $\hat{q}_4$ and $\hat{v}_4$ since $\mu_r$ and $\sigma_{rr}$ are not required.

Rates of convergence are summarized in Proposition 1 below. The proof of this result follows easily from Van Allen *et al.* (2008) and the development above.

**Proposition 1.** Assume a discrete network satisfying (3) and let $m \to \infty$. The query mean $\mu_q$ remains constant while the variance $\sigma_{qq}$ approaches zero at rate $m^{-1}$. The mean approximations have errors $\hat{q}_j - \mu_q$ approaching zero at rate $m^{-1}$ for $j = 1$ and 2, and at the faster rate $m^{-3/2}$ for $j = 3$ and 4. All four variance approximations have relative errors $(\hat{v}_j - \sigma_{qq})/\sigma_{qq}$ approaching zero at rate $m^{-1}$.

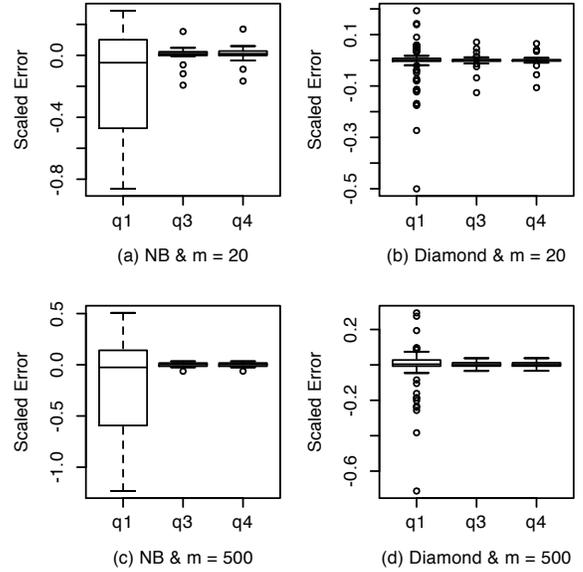

Figure 2: Boxplots of scaled errors $m(\hat{q}_j - \hat{q}_0)$ for $j \in \{1, 3, 4\}$, $m \in \{20, 500\}$, and network structures NB and Diamond. Each boxplot shows variation in errors for a set of distinct queries, $2^2 + 2^4 = 20$ for NB and 108 for Diamond. Errors for $\hat{q}_3$ and $\hat{q}_4$ are nearly identical. Errors for $\hat{q}_1$ are often much larger. Results for $\hat{q}_2$ are not plotted since $\hat{q}_2 - \hat{q}_0 \approx 2(\hat{q}_1 - \hat{q}_0)$.

## 5 NUMERICAL RESULTS

We evaluated accuracy of approximations $\hat{q}_j$ and $\hat{v}_j$ using highly accurate empirical estimates of $\mu_q$ and $\sigma_{qq}$. These estimates $\hat{q}_0$ and $\hat{v}_0$ were obtained by simulating $k = 10^6$ replicates of $\boldsymbol{\Theta}$ from the posterior distribution, evaluating $q(\boldsymbol{\Theta})$ for each replicate, then calculating the sample mean and sample variance. Computational costs preclude using empirical variance estimates in practice. When $m$ is large, asymptotic normality of $q(\boldsymbol{\Theta})$ implies that the distribution of $\hat{v}_0/\sigma_{qq}$ is approximately $(1/k)\chi_k^2$ with variance $2/k$. Consequently $\hat{v}_0/\sigma_{qq}$ varies over the interval $1 \pm 2\sqrt{2/k}$ for roughly 95% of samples. Since our variance approximations have relative errors of order $m^{-1}$, it follows that $k$ should be of order at least $m^2$ for $\hat{v}_0$ to have substantially smaller relative error. When comparing approximate relative errors $(\hat{v}_j - \hat{v}_0)/\hat{v}_0$ with $k = 10^6$, variation in $\hat{v}_0$ has a noticeable effect for $m = 500$; see Figure 3(f).

Our examples differ with respect to network structure, posterior distribution, and query. All variables are binary. All posterior distributions satisfy BDe constraints (e.g., see Hooper 2008), so all variables have the same effective sample size $m$. Hyperparameters are thus determined by $m$ and the poste-rior means $E\{\Theta \mid \mathcal{D}\}$. Our examples are from three small networks, each with one vector $E\{\Theta \mid \mathcal{D}\}$ and $m \in \{20, 50, 100, 200, 500\}$:

- Two naïve Bayes networks (NB-2 and NB-4 with 2 and 4 features plus the root variable); $H$ = root,

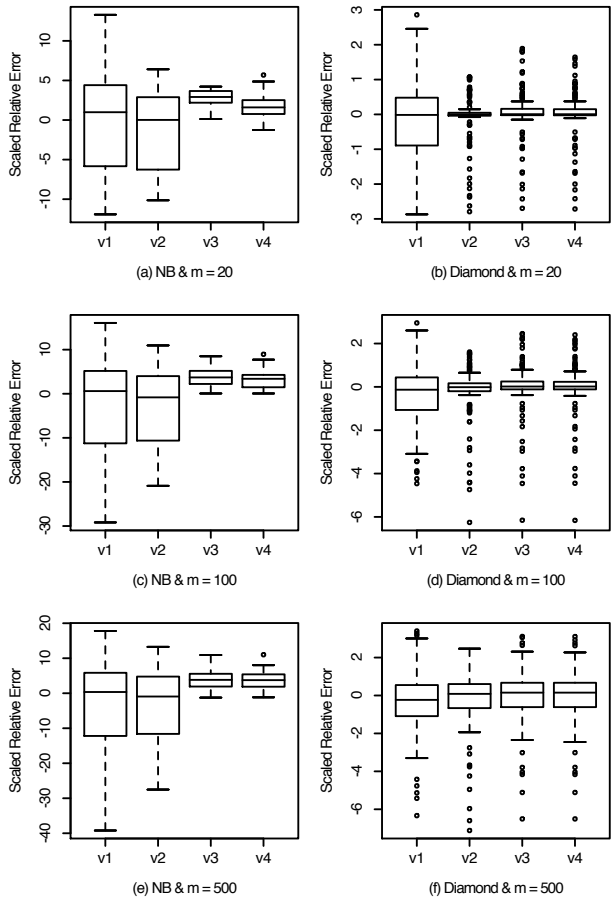

Figure 3: Boxplots of relative errors $m(\hat{v}_j - \hat{v}_0)/\hat{v}_0$ for $j \in \{1, 2, 3, 4\}$, $m \in \{20, 100, 500\}$, and network structures NB and Diamond. Each boxplot shows variation among values for a set of distinct queries, 20 for NB and 108 for Diamond. We observe that: relative errors tend to be larger for NB compared with Diamond; $\hat{v}_3$ and $\hat{v}_4$ tend to over-estimate $\sigma_{qq}$ for NB and are more accurate than $\hat{v}_2$; the three methods $\hat{v}_2$, $\hat{v}_3$, and $\hat{v}_4$ have similar accuracy for Diamond; $\hat{v}_1$ is less accurate than the other methods. The four methods appear to have similar accuracy in (f), but these plots are misleading. Many of the Diamond queries have the property described following (12), where $\hat{v}_2 = \hat{v}_3 = \hat{v}_4 = \sigma_{qq}$. We would therefore expect the Diamond results for $m = 500$ to be similar to those for $m = 100$. It appears that the variation among relative errors for $m = 500$ is due in large part to variation in $\hat{v}_0$.

$E$ = all children of $H$, $e$ varies over all combinations ($2^2$ for NB-2, $2^4$ for NB-4).

- Diamond network with 4 variables ◇, all 108 distinct queries with one hypothesis variable.

Approximations for means are compared in Figure 2 and for variances in Figure 3. The errors and relative errors are multiplied by $m$ in these figures to facilitate comparisons across a range of effective sample sizes. Boxplots for $m = 20$, 100, and 500 are shown. Plots for other values of $m$ are similar. By Proposition 1, relative errors $(\hat{v}_j - \sigma_{qq})/\sigma_{qq}$ should approach zero at rates $c_j/m$, where $c_j$ depends implicitly on the network, $E(\Theta \mid \mathcal{D})$, and the query. This theory is supported by Figure 3 and additional plots (not shown) comparing the four methods for individual queries. Our results suggest that $c_3 \approx c_4$ while $c_1$ and $c_2$ tend to be further from zero. Relative errors can be interpreted in terms of variances or standard deviations. If $(\hat{v}_j - \sigma_{qq})/\sigma_{qq} = c_j/m$, then we have

$$\frac{\hat{v}_j}{\sigma_{qq}} = 1 + \frac{c_j}{m} \quad \text{and} \quad \frac{\sqrt{\hat{v}_j}}{\sqrt{\sigma_{qq}}} = \sqrt{1 + \frac{c_j}{m}} \approx 1 + \frac{c_j}{2m}.$$

## 6 COMPUTATIONAL ISSUES

Inference in Bayesian networks is in general an NP-complete problem (Cooper, 1990). For instance, the complexity of the Variable Elimination (VE) Algorithm is $O(d^r)$, where $d$ is an upper bound on the number of values that a variable can take and $r$ is an upper bound on the size of a factor generated by the VE Algorithm (Koller and Friedman, 2008). Network doubling uses essentially the same technique to calculate a variance as that used to evaluate a query, resulting in corresponding computational complexity. The doubled CPtables are larger (squared number of rows and columns), so the computational complexity of VE is increased to $O(d^{2r})$. The delta method retains $O(d^r)$ complexity (Van Allen *et al.*, 2008), so is typically faster in large networks; see Table 2 below.

In some cases, we can exploit the structure of the network or query to achieve a polynomial time inference algorithm. For poly-tree Bayesian networks (i.e. networks with at most one undirected path between any pair of nodes), there exist inference algorithms with linear time complexity. Reduced complexity is also available when the query can be expressed in terms of probabilities of hypothesis and evidence nodes conditioned on their Markov blanket; i.e., the parents, the children and the parents of the children. Once again, we have a polynomial time inference algorithm. These techniques translate directly to efficient algorithms for computing all of the variance approximations in Table 1.

Table 2: Timing results in seconds.

| Network | # Queries | Delta | Doubling |
|---|---|---|---|
| NB-4 | 100,000 | 37.837 | 3.969 |
| Diamond | 108,000 | 50.052 | 12.660 |
| Alarm | 100 | 11.390 | 282.342 |

We empirically compared timing results for the delta and network doubling methods using queries from three networks: 1000 from Naïve Bayes (5 variables) repeated 100 times, 108 from the Diamond network (4 variables) repeated 1000 times, and 100 from the Alarm network (37 variables, Beinlich *et al.* 1989). The Alarm queries were randomly generated so that queries had on average 3 hypothesis variables, 25 evidence variables, and 9 non-specified variables. The results in Table 2 corroborate the earlier claim that doubling is faster for simple queries and delta is faster for complex queries.

## 7 CONCLUSION

We plan to extend the implementation of the network doubling and delta methods to continuous and hybrid networks. This should be easy for the doubling method when all continuous variables are evidence variables. We will then compare the accuracies of these methods.

Our main contributions are summarized as follows.

- Development of a network doubling method to approximate query variances. This technique exploits the Cooper and Herskovits formula (5) for approximating the query mean and is easily implemented for discrete networks. The technique is also applicable for continuous and hybrid networks, but implementation may be less straightforward.

- Adjustments to improve accuracy, motivated by asymptotic theory for discrete networks. This theory also provides rates of convergence for the approximations.

- Numerical comparisons of the network doubling and delta methods, showing superior accuracy of the former in simple networks.

We do not recommend that network doubling replace delta in all applications. If the effective sample size is large, then both approaches may provide adequate approximations and the choice between them may depend primarily on computational complexity. If the network is large, then delta may have an advantage. If the effective sample size is small or the network is not large, then doubling may be the better choice.

**Acknowledgements**

We are grateful for helpful comments from the anonymous reviewers. We acknowledge support provided by NSERC, iCORE, and the Alberta Ingenuity Centre for Machine Learning.

**References**

I. Beinlich, H. Suermondt, R. Chavez, and G. Cooper (1989). The ALARM monitoring system: A case study with two probabilistic inference techniques for belief networks. In *Second European Conference on Artificial Intelligence in Medicine*, **38**: 247–256.

G. Cooper (1990). The computational complexity of probabilistic inference using Bayesian belief networks. *Artificial Intelligence* **42**: 393–405.

G. Cooper and E. Herskovits (1991). A Bayesian method for the induction of probabilistic networks from data (Report SMI-9M). Medical Informatics, Univ. of Pittsburgh. (Also available as Report KSL-91-02, Stanford Univ., Medical Informatics.)

G. Cooper and E. Herskovits (1992). A Bayesian method for the induction of probabilistic networks from data. *Machine Learning* **9**: 309–347.

A. Gelman, J. Carlin, H. Stern and D. Rubin (2003). *Bayesian Data Analysis, Second Edition*. New York: Chapman and Hall.

Y. Guo and R. Greiner (2005). Discriminative model selection for belief net structures. In *Proceedings of the Twentieth National Conference on Artificial Intelligence (AAAI-05)*, 770-776.

P. M. Hooper (2008). Exact distribution theory for belief net responses. *Bayesian Analysis* **3**: 615–624.

N. L. Johnson and S. Kotz (1972). *Distributions in Statistics: Continuous Multivariate Distributions*. New York: John Wiley & Sons.

D. Koller and N. Friedman (2008). *Structured Probabilistic Models*, in preparation.

C. Lee, R. Greiner, and S. Wang (2006). Using query-specific variance estimates to combine Bayesian classifiers. In *Proceedings of the International Conference on Machine Learning (ICML-06)*, 529-536.

T. Van Allen, R. Greiner, and P. M. Hooper (2001). Bayesian error-bars for belief net inference. In *Proceedings of the Seventeenth Conference on Uncertainty in Artificial Intelligence (UAI-01)*, 522-529.

T. Van Allen, A. Singh, R. Greiner, and P. Hooper (2008). Quantifying the uncertainty of a belief net response: Bayesian error-bars for belief net inference. *Artificial Intelligence* **172**: 483–513.